\documentclass{article}

\usepackage{microtype}
\usepackage{amsmath}

\usepackage{amssymb}
\usepackage{gensymb}
\usepackage{graphicx}
\usepackage{pgf}
\usepackage{float}
\usepackage{ifdraft}
\usepackage[hyphens]{url}
\usepackage{hyperref}
\usepackage{enumitem}

\usepackage{eqexpl}
\eqexplSetIntro{where:} % set parenthesis in the left of the first item
\eqexplSetDelim{=} % set delimiter to "="

\usepackage{multicol}

\usepackage{siunitx}
\usepackage{tabularx}
\usepackage{booktabs}
\usepackage{multirow}
\usepackage{tablefootnote}
\usepackage{tabularray}
\UseTblrLibrary{booktabs}
\UseTblrLibrary{counter}

\usepackage{tikz}
\usetikzlibrary{positioning}
\usetikzlibrary{shapes.geometric}
\usetikzlibrary{shapes.arrows}
\usetikzlibrary{decorations.pathmorphing, decorations.pathreplacing, calc}

\usepackage{tikz-cd}
\tikzcdset{
    math mode=false
}

\usepackage{titlesec}

\usepackage{caption}
\usepackage[numbers,sort&compress]{natbib}

\usepackage{adjustbox}

\usepackage[titletoc,title]{appendix}

\usepackage{svg}

\usepackage{subcaption}

\usepackage{afterpage}
\usepackage[section]{placeins}

\usepackage{nth}

\usepackage{xstring}

\usepackage{authblk}

\usepackage{todonotes}
\setuptodonotes{inline,backgroundcolor=yellow!20,prepend,caption={TODO}}

\title{Accelerating Transient CFD through Machine Learning-Based Flow Initialization}

\author[1]{Peter Sharpe}
\author[1]{Rishikesh Ranade}
\author[1]{Kaustubh Tangsali}
\author[1]{Mohammad Amin Nabian}
\author[1]{Ram Cherukuri}
\author[1]{Sanjay Choudhry}
\affil[1]{NVIDIA Corporation}
\date{\today}

\begin{document}
\maketitle

\begin{abstract}
    Transient computational fluid dynamics (CFD) simulations are essential for many industrial applications, but suffer from high compute costs relative to steady-state simulations. This is due to the need to: (a) reach statistical steadiness by physically advecting errors in the initial field sufficiently far downstream, and (b) gather a sufficient sample of fluctuating flow data to estimate time-averaged quantities of interest. We present a machine learning-based initialization method that aims to reduce the cost of transient solve by providing more accurate initial fields. Through a case study in automotive aerodynamics on a 17M-cell unsteady incompressible RANS simulation, we evaluate three proposed ML-based initialization strategies against existing methods. Here, we demonstrate 50\% reductions in time-to-convergence compared to traditional uniform and potential flow-based initializations. Two ML-based initialization strategies are recommended for general use: (1) a hybrid method combining ML predictions with potential flow solutions, and (2) an approach integrating ML predictions with uniform flow. Both strategies enable CFD solvers to achieve convergence times comparable to computationally-expensive steady RANS initializations, while requiring far less wall-clock time to compute the initialization field. Notably, these improvements are achieved using an ML model trained on a different dataset of diverse automotive geometries, demonstrating generalization capabilities relevant to specific industrial application areas. Because this Hybrid-ML workflow only modifies the inputs to an existing CFD solver, rather than modifying the solver itself, it can be applied to existing CFD workflows with relatively minimal changes; this provides a practical approach to accelerating industrial CFD simulations using existing ML surrogate models.
\end{abstract}

\section{Introduction}

As computational resources continue to grow, computational fluid dynamics (CFD) simulations have been brought in earlier and earlier into the engineering design and analysis process -- supplementing tools like hand calculations and experimental testing in the engineer's toolbox. Historically, many CFD simulations of industrial relevance have been formulated as steady-state problems, where the time derivatives in the governing equations are set to zero during discretization, and turbulence (if applicable) is handled through Reynolds-averaging of the governing Navier-Stokes equations. For problems that readily admit steady-state solutions (e.g., fully-attached flows, with time-invariant boundary conditions), this approach typically requires only a fraction of the computational resources required for the alternative: a transient or pseudo-transient problem that is solved via time-marching to obtain a time-averaged solution.

However, some industrial CFD applications do not allow sufficient practical engineering insight to be extracted if approximated as steady-state simulations. This is most obvious in cases where the physics themselves are \textit{fundamentally} time-dependent, such as the massively-separated flow in the wake of a bluff body. On such problems, a steady-state discretization often will either fail to converge, or will converge to a very different solution than a transient simulation due to an inability to capture important flow instabilities\footnote{This is particularly true in cases where large, coherent vortex structures are shed, such as in classic high-Reynolds-number cylinder flow simulations.} \cite{benimModellingTurbulentFlow2008}. In such cases, a transient discretization of the governing equations is required to achieve accurate results on engineering quantities of interest.

Even on problems where steady-state solutions are computable and useful, growing computational resources have enabled increasing adoption of transient formulations that can capture previously-unresolved flow physics. For these \emph{fundamentally-transient} models, and in the case of high Reynolds numbers, no well-posed steady-state variant exists. For example, large-eddy simulations (LES) and delayed detached-eddy simulations (DDES) are now commonplace in the automotive \cite{hupertz2022autocfd, ashton2018assessing, ashtonDrivAerMLHighFidelityComputational2024} and aerospace \cite{clarkHighLiftPredictionWorkshop2025} industries, and direct numerical simulations (DNS) are occasionally found in academic research on turbulence. Likewise, lattice-Boltzmann methods (LBM) have received renewed attention due to their compatibility with hardware accelerators and parallel computing. All of these methods allow unsteady flow physics (like turbulence) to be directly resolved to varying degrees, potentially resulting in more accurate simulations in cases where the assumptions of Reynolds-averaging and turbulence closure modeling are violated.

In a typical transient CFD workflow for industrial applications, the choice of field initialization strategy has a significant impact on the overall computational cost. This is because any inaccuracies in the initial field must sufficiently dampen or advect downstream before meaningful time-averaging of a statistically-steady solution can begin. To conceptually illustrate this, consider a localized error in total pressure within an incompressible flow field. The total pressure itself can be expressed as an advected quantity with minimal diffusion, and so the error will propagate downstream at the local flow velocity. However, this local error will cause \emph{global} changes in the flow field, due to the elliptic nature of the pressure Poisson equation. This global error kernel will diminish with distance to the source of the error (roughly in accordance with a Green's function to the Poisson equation), and hence a statistically-steady solution can only be reached once the error has \emph{physically} propagated sufficiently far downstream.

Arguably, the choice of initialization is often more important in transient cases than in a steady-state case. This is because steady-state solvers can use \emph{numerical} techniques (e.g., geometric-algebraic multigrid, implicit timestepping with large CFL numbers\footnote{While high CFL numbers can also be used in transient cases to help ``wash'' the initial field, this application is much more stability-limited than the steady case, as the large nonzero time derivative term acts to amplify numerical instabilities.}) to quickly correct errors in the initial field. Conversely, transient solvers (which have physically-meaningful intermediate solutions) rely on \emph{physical} damping or advection to correct errors in the initial field. Therefore, a transient solve is limited by the speed of physical information propagation. In high-Reynolds-number flows (whether incompressible or compressible), the rate-limiting physical transport processes concern advected quantities like vorticity or total pressure; these are advected downstream at the local flow velocity. This makes inaccuracies in the initial field more costly in transient cases; reaching statistical steadiness often requires at least one ``flow-through time'' in internal flow cases, or 40-80 ``convective time units'' (CTUs)\footnote{A CTU is defined as $\text{CTU} = |\vec{U}_\infty|/l_0$, where $\vec{U}_\infty$ is the free-stream velocity and $l_0$ is a characteristic streamwise length of the major flow features, such as the wheelbase of a car.} in external flow cases \cite{rumseyHighLiftPredictionWorkshops2024}.

Traditionally, many transient CFD simulations are initialized either with a simple uniform field\footnote{Often with values taken from the boundary conditions}, or with an approximate solution from a steady-state solver (e.g., a steady RANS solver). Although these initializations from steady-state typically reduce the cost of the subsequent transient solve, the steady-state solution itself can take a significant amount of time to compute.

In this work, we present an alternative initialization strategy using approximate solutions from machine learning surrogate models. These surrogate models can be evaluated in seconds, and they can be pre-trained to generalize across an industrially-relevant range of cases and flow conditions\footnote{To illustrate the magnitude of generalization described here: the DoMINO ML surrogate model used in this work is currently capable of reasonable generalization across a range of automotive geometries (e.g., both pickup trucks and sports cars) and freestream velocities. This same DoMINO model currently would likely not, however, generalize to automotive cases with flow physics that differ significantly from the training dataset (e.g., nonzero sideslip angles, multiple cars in proximity, etc.), or to cases studying aircraft or buildings.}. We show that using these surrogate models as initializations can reduce the time required to reach a statistically-steady solution by 50\% or more, offering a significant speedup for transient CFD simulations that integrates seamlessly with existing CFD workflows.

The key contributions of this work are:
\begin{enumerate}
    \item Implementation and validation of two ML-based initialization strategies that reduce time-to-convergence by 50\% while requiring only seconds of computation time, with generalization capabilities that enable applicability beyond the training dataset
    \item A comparison of initialization strategies for transient CFD, including traditional approaches (uniform flow, potential flow, steady RANS, and DDES) and novel ML-based methods
\end{enumerate}

\section{Methods}

\subsection{Case Study: Automotive Aerodynamics}

\begin{figure}[h]
    \centering
    \includegraphics[width=\textwidth]{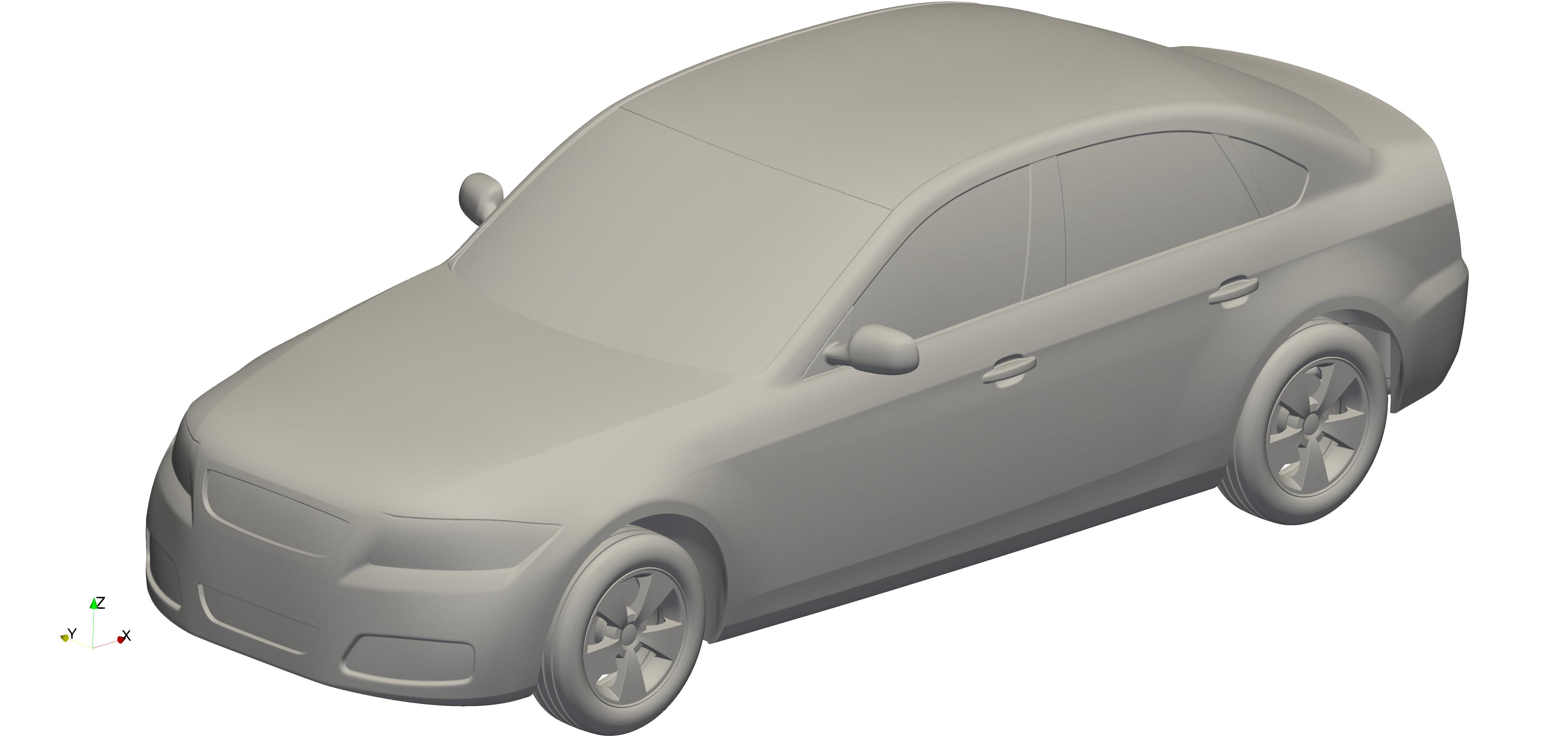}
    \caption{Vehicle geometry used in the case study, from run number 4 of the DrivAerML dataset.}
    \label{fig:vehicle}
\end{figure}

The case study for this work is an external aerodynamics flow over a sedan, with the vehicle geometry shown in Figure \ref{fig:vehicle}. This case study is reproduced from the publicly-available DrivAerML high-fidelity dataset by Ashton et al. \cite{ashtonDrivAerMLHighFidelityComputational2024}, which contains 500 simulations of varying vehicle designs to support the development of machine learning surrogates for automotive aerodynamics. The case used in the current study is run number 4 from the DrivAerML dataset. This case was chosen because its drag force was near the median of values in the dataset (using drag values computed by Ashton et al.); this is intended to provide a representative example of an automotive aerodynamics case.

\subsubsection{Geometry and Flow Model}

The flow is modeled as incompressible and transient (unsteady), with a density of $\rho = 1.225 \rm\ kg/m^3$ and a kinematic viscosity of $\nu = 1.507 \times 10^{-5} \rm\ m^2/s$. A $k$-$\omega$ SST turbulence model \cite{menterTwoequationEddyviscosityTurbulence1994} is used: this is the main notable difference from the original case study by Ashton et al. \cite{ashtonDrivAerMLHighFidelityComputational2024}, which used delayed detached-eddy simulation (DDES) modeling approach\footnote{The DDES formulation in this case uses a Spalart-Allmaras RANS model \cite{spalartOneequationTurbulenceModel1992} in the near-wall region, and a LES model in the outer region and in separated flow.}. This unsteady RANS (URANS) formulation was chosen to allow grid-independence to be achieved at much coarser mesh resolutions than a DDES formulation \cite{balin2018comparison}, enabling much faster experimentation. More precisely, URANS is the cheapest well-known modeling approach that allows the central difficulty of transient CFD convergence to be captured: namely, the need to physically advect error sufficiently far downstream. Within URANS, the choice of a $k$-$\omega$ SST closure over other RANS turbulence models was motivated by its well-documented ability to handle adverse pressure gradients and flow separation in external aerodynamics \cite{menterTwoequationEddyviscosityTurbulence1994}.

The fluid domain is a rectangular prism extending 40 meters upstream and downstream and 22 meters to either side of the datum, which is the center of the vehicle's front axle. The domain is extended 20 meters in height above the road plane. 

The inlet is a standard velocity-inlet boundary condition, with Dirichlet values of freestream velocity $U_\infty = 38.889 \rm\ m/s$, turbulent kinetic energy of $k = 0.24 \rm\ m^2/s^2$, and specific dissipation rate of $\omega = 1.78 \rm\ s^{-1}$. The outlet is a standard pressure-outlet boundary condition, with a static gauge pressure of $p = 0 \rm\ Pa$. The side and top boundaries are modeled as slip walls. The bottom boundary is a slip wall upstream of $2.339 \rm\ m$ forward of the datum, and a no-slip wall aft of this point. All no-slip walls (i.e., the vehicle surface and the aft half of the bottom boundary) use a $k$-based wall function that computes first-node eddy viscosity based on the log-law of the wall.

\subsubsection{Mesh}

\begin{figure}[h]
    \centering
    \includegraphics[width=\textwidth]{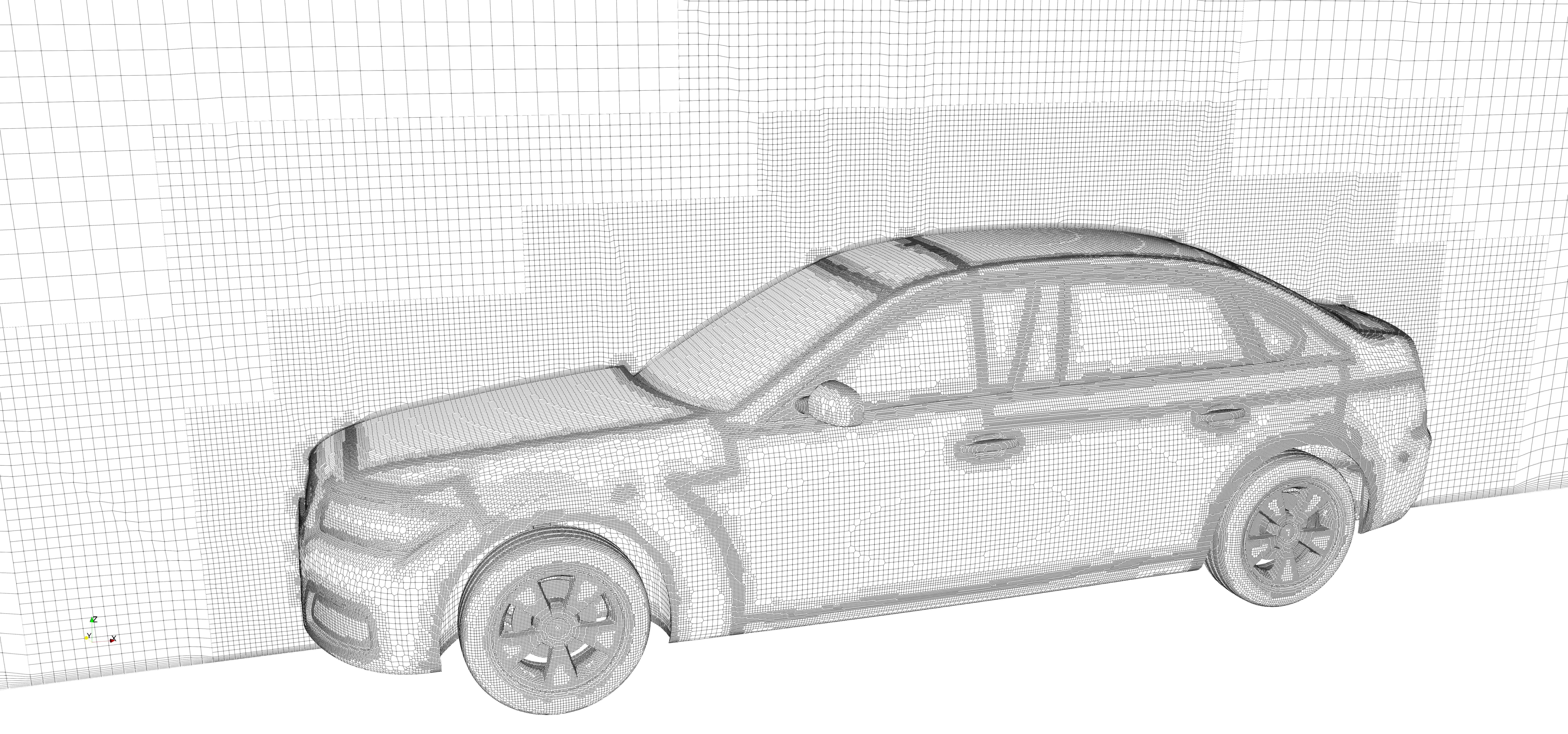}
    \caption{Mesh used in the case study, visualized on the vehicle surface and centerline plane.}
    \label{fig:mesh}
\end{figure}

The mesh, shown in Figure \ref{fig:mesh}, is hexahedral-dominated and generated with the OpenFOAM-based snappyHexMesh tool. The mesh contains 16.7 million cells, a value that was chosen on the basis of a grid-independence study that also included meshes with approximately 8 million and 27 million cells. A boundary layer mesh is generated on all no-slip walls with a median $y^+$ of 36, fitting comfortably within the log-law region assumed by the wall function approach\footnote{The \nth{5} and \nth{95} percentiles of $y^+$ values are 7.4 and 137, respectively. Values are computed using outer-layer velocity values obtained from subsequent solutions.}.

\subsubsection{CFD Solver}

The transient flow solver used is OpenFOAM v2206, using the corrected PIMPLE algorithm for segregated pressure-velocity coupling. The flow is time-integrated using an Euler scheme and a fixed timestep of $\Delta t = 2 \times 10^{-4} \rm\ s$, corresponding to a mean Courant number of 0.015 and a maximum Courant number of 58.

Two corrector sub-iterations are used for the pressure solve on each timestep, which was necessary to stabilize the time integration for cases where initializations had high errors -- specifically, for the uniform flow case, and to a lesser extent, the potential flow case.

Spatial schemes are mixed-order, to emphasize robustness towards various initialization strategies. Convective, diffusive, and gradient terms are second-order\footnote{TVD flux limiting is used, which drops the order of accuracy near discontinuities. This is useful during the initial transient, though final results are smooth and hence second-order.} accurate for momentum and pressure, and first-order accurate for turbulence quantities using upwinding.

Simulations were run using 40 CPU cores on a dual-socket E5-2698 v4 server for 2 seconds of physical simulation time, requiring approximately 51 hours of wall-clock time per simulation. This physical simulation time is equivalent to 27.0 convective time units (CTUs), where a CTU is defined as $\text{CTU} = |\vec{U}_\infty|/l_0$, with $\vec{U}_\infty=38.889 \rm\ m/s$ the freestream velocity and $l_0=2.88 \rm\ m$ the vehicle's wheelbase.

\subsection{Initialization Strategies}
\label{sec:init_strategies}

\subsubsection{Traditional Strategies}
\label{sec:init_strategies_traditional}

In the numerical experiment conducted here, the field values on the mesh (which consist of the velocity $\vec{U}$, the pressure $p$, and the turbulence quantities $k$ and $\omega$) were initialized using one of several different strategies. The traditional strategies that were tested are:

\begin{itemize}
    \item \textbf{Uniform Flow}: Values are taken directly from the boundary conditions, and applied everywhere: $\vec{U} = (U_\infty, 0, 0)$, $p = 0$, $k = 0.24 \rm\ m^2/s^2$, and $\omega = 1.78 \rm\ s^{-1}$.
    \item \textbf{Potential Flow}: The $\vec{U}$ and $p$ fields are taken from a potential flow solution performed on the same mesh. The $k$ and $\omega$ fields are given a uniform initialization (using the same values as the uniform flow case), since potential flow theory assumes inviscid, irrotational flow and therefore does not model turbulence quantities. This potential flow solution required a wall-clock execution time of 11 minutes on the 40-core machine used throughout this study.
    \item \textbf{Steady-State RANS Flow}: All fields are taken from a steady-state RANS solution performed on the same mesh. Identical settings are used, with the exception of replacing the pressure-velocity coupling with a SIMPLE algorithm. The RANS solution itself is initialized with the potential flow solution. Though the RANS solution never reaches a true steady-state, statistical steadiness (as measured by the procedure later described in Section \ref{sec:convergence_metric}) is achieved after 477 iterations, corresponding to a wall-clock execution time of 2.4 hours.
    \item \textbf{DDES Flow}: All fields are initialized using the time-averaged fields from a DDES simulation of the same case, performed by Ashton et al. \cite{ashtonDrivAerMLHighFidelityComputational2024}. This DDES solution was performed on a different mesh with 137 million cells, so the solutions are interpolated to the current mesh using inverse distance weighting from cell centers. Notably, because the DDES solution does not use a $k$-$\omega$ turbulence model, the $k$ and $\omega$ fields are initialized to uniform values. The authors acknowledge that initializing a URANS transient solve with a DDES solution would be quite atypical in industrial practice. The main motive for its inclusion in this study is instead more theoretical: we aim to assess how transient convergence time is affected by initializing from a solution with similar turbulence modeling (i.e., RANS $\rightarrow$ URANS) vs. an initialization with fundamentally-different turbulence modeling (DDES $\rightarrow$ URANS). 
\end{itemize}

\subsubsection{ML-based Strategies}
\label{sec:init_strategies_ml}
In addition, several machine learning-based initialization strategies were tested, all leveraging the DoMINO architecture developed by Ranade et al. \cite{ranadeDoMINODecomposableMultiscale2025} through different integration approaches. DoMINO (Decomposable Multi-scale Iterative Neural Operator) is a neural operator architecture within the NVIDIA PhysicsNeMo framework that learns geometric encodings from point cloud data to predict PDE solutions. The model operates on discrete domain points by dynamically constructing numerical stencils from local neighborhood information, enabling simultaneous prediction of flow quantities on both geometric surfaces and within the surrounding fluid volume. This dual prediction capability is particularly critical for applications like automotive aerodynamics where both surface quantities (e.g., pressure distributions) and volumetric features (e.g., wake structures) inform key design decisions.

The architecture employs a multi-scale approach that captures local flow features through hierarchical geometric processing while maintaining global consistency through iterative solution refinement. Complete architectural details are provided in \cite{ranadeDoMINODecomposableMultiscale2025}, with an open-source implementation available in the PhysicsNeMo repository \cite{physicsnemo}.

A DoMINO surrogate model was trained on the NVIDIA-internal DriveSim dataset, which contains cases of steady-state RANS solutions for automotive aerodynamics problems. The dataset consists of 1,000 geometrically morphed variants of different vehicle classes (sedans, SUVs, hatchbacks, pickups, vans, etc.) simulated at speeds ranging from 20 to 50 m/s.
While both DriveSim and DrivAerML are automotive aerodynamics datasets, the geometries used in each are generated using different morphing procedures, and start with different basic geometries. Likewise, there are differences in boundary conditions and flow physics between the two datasets. For example, DriveSim varies the inlet velocity (contrasted with DrivAerML's fixed velocity), and kinematic viscosity values are different. Therefore, inference using this DoMINO model on this case provides a reasonable proxy for real-world transfer learning, where a model trained on a single dataset is used to initialize a somewhat-similar, but not identical, problem.

\begin{figure}[h]
    \centering
    \begin{subfigure}{\textwidth}
        \centering
        \includegraphics[width=\textwidth]{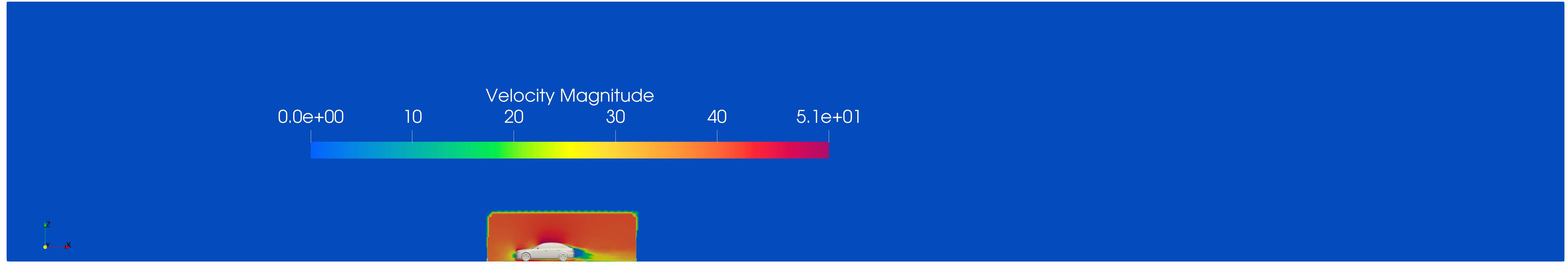}
        \caption{Full domain view showing the velocity magnitude predicted by the DoMINO surrogate model.}
        \label{fig:ml_init_u_full}
    \end{subfigure}
    \begin{subfigure}{\textwidth}
        \centering
        \includegraphics[width=\textwidth]{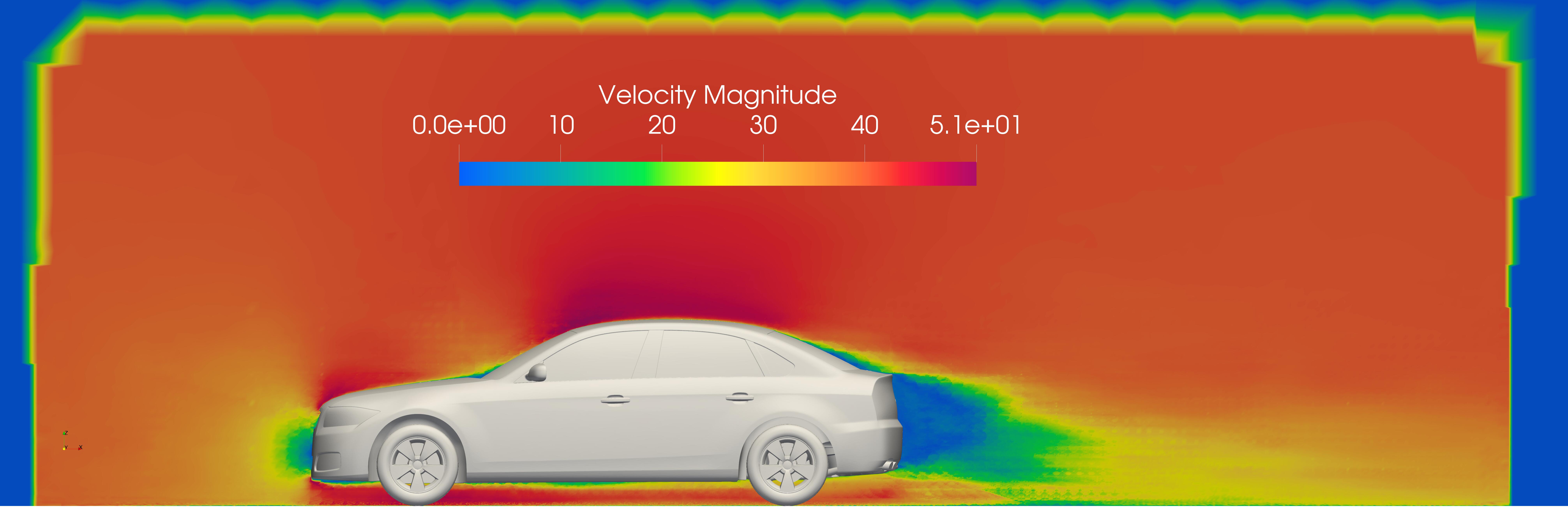}
        \caption{Zoomed view of the near-field region, showing velocity magnitude predicted by the DoMINO surrogate model.}
        \label{fig:ml_init_u_zoom}
    \end{subfigure}
    \caption{Velocity field predictions from the DoMINO surrogate model. As DoMINO is a point-cloud-based model, the resulting point data is interpolated to cells for visualization purposes. Note the large regions with a predicted velocity of zero, showing where the CFD domain extends beyond DoMINO's evaluation domain; here, the DoMINO solution must be extended through one of the various strategies described in the text.}
    \label{fig:ml_init_u}
\end{figure}

For balancing the tradeoff between computational efficiency and model accuracy, the DoMINO surrogate model is trained and evaluated in a bounding box constructed around the vehicle, as shown by the velocity field in Figure \ref{fig:ml_init_u}. This represents a near-field region of the domain that consists of the most relevant flow structures that have the highest impact on the aerodynamic forces exerted on the vehicle. This therefore leaves various possible strategies to extend the DoMINO solution to the full domain. In one tested initialization approach, uniform flow values are used for the far-field region, while in another approach, the DoMINO solution is extended to the full domain using an inverse distance weighting (IDW) approach, where extra points are added to the boundary of the full domain based on known boundary condition values.

In a third approach, the DoMINO solution is combined with the potential flow solution described in Section \ref{sec:init_strategies_traditional}: DoMINO is used for boundary layers and wakes, while the potential flow solution is used for the rest of the domain. This is intended to play on the strengths of both approaches. The potential flow solution has the advantage of satisfying conservation laws, which results in fewer numerical artifacts than the DoMINO solution in regions with subtle field differences, like the far-field. Moreover, potential flow should be a very good initialization in the far-field, where vorticity is near-zero. On the other hand, potential flow is wholly unable to capture flow physics within regions with vorticity, causing very large error in boundary layers and wakes\footnote{In addition, sharp exterior corners theoretically result in infinite flow velocity in potential flow, which is non-physical and can lead to numerical instabilities.}. Conveniently, these near-field regions are exactly where DoMINO is most effective.

\begin{figure}[h]
    \centering
    \includegraphics[width=\textwidth]{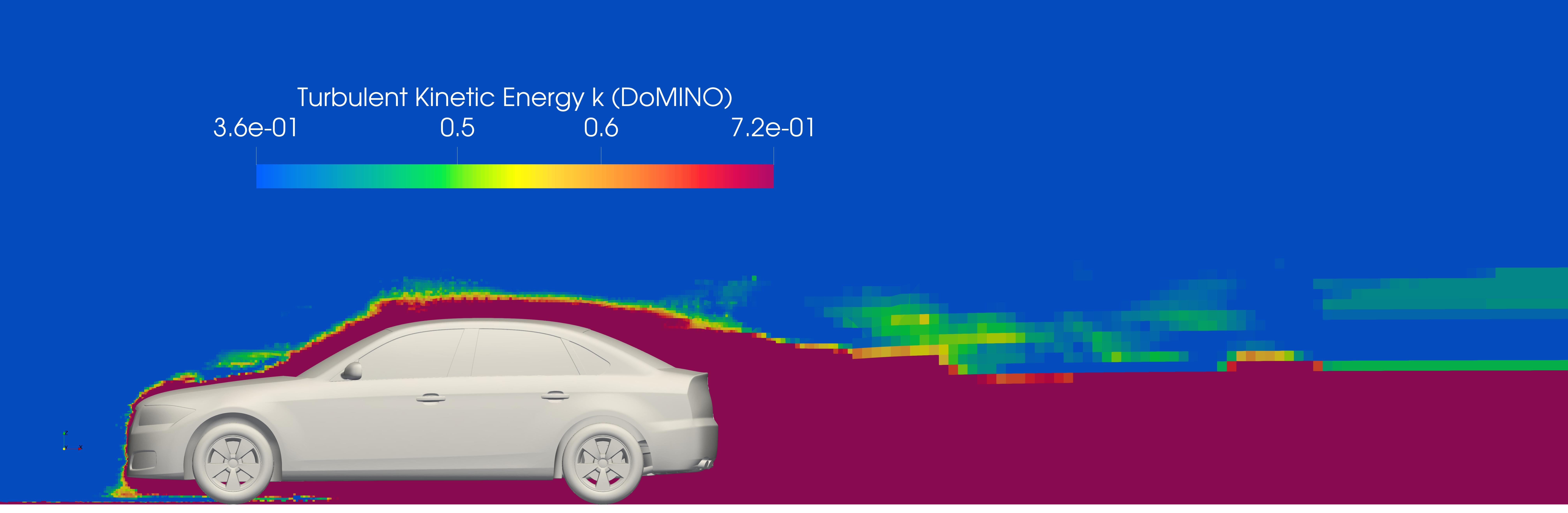}
    \caption{Visualization of active regions for the $k$-based intermittency function used to smoothly blend the DoMINO and potential flow solutions. Red regions use the DoMINO solution, while blue regions use the potential flow solution. $k$ field predicted by DoMINO.}
    \label{fig:ml_init_alpha}
\end{figure}

In order to merge the DoMINO solution with the potential flow solution, the DoMINO solution is first extended to the full domain using the previously-described IDW approach. Then, the DoMINO-predicted values for turbulent kinetic energy $k$ are used to blend the DoMINO solution and potential flow solutions together. This strategy is based on theoretical properties of the governing PDE for $k$ in the $k$-$\omega$ SST turbulence model \cite{menterTwoequationEddyviscosityTurbulence1994} used for training data. Specifically, $k$ is produced only in regions of high vorticity, and is advected downstream once produced. Far-field regions not only produce minimal $k$, but instead are typically net sinks of $k$ due to decay. Therefore, high $k$ is a natural indicator (across a wide range of external flow cases) for where the underlying assumptions of potential flow solutions are invalid, and hence where the DoMINO prediction should be preferred. Based on visual inspection, it was found that an isosurface of $k = 2 k_\infty$ consistently isolates the regions with near-zero vorticity, where $k_\infty$ is the inlet turbulent kinetic energy. Therefore, the two solutions were blended as follows:
\begin{equation}
    k_\infty = \SI{0.24}{\meter\squared\per\second\squared}, \qquad k_\text{lower} = 1.5k_\infty, \qquad k_\text{upper} = 3k_\infty
\end{equation}

\begin{equation}
    \alpha = \sin^2\left(\frac{\pi}{2} \cdot \text{clip}\left(\frac{k_\text{DoMINO} - k_\text{lower}}{k_\text{upper} - k_\text{lower}}, 0, 1\right)\right)
\end{equation}

\begin{equation}
    \phi = \alpha\ \phi_\text{DoMINO} + (1-\alpha)\ \phi_\text{PF}
\end{equation}

\noindent where $\phi$ represents any field value, $\phi_\text{DoMINO}$ and $\phi_\text{PF}$ are the corresponding values from the DoMINO and potential flow solutions respectively, and $\alpha$ is a weighting parameter that varies smoothly from 0 to 1. The clip function restricts its argument to the interval $[0,1]$. The resulting intermittency function $\alpha$ can be visualized in Figure \ref{fig:ml_init_alpha}, where red regions ($k_\text{DoMINO} > k_\text{upper} = 0.72 \rm\ m^2/s^2$) use the DoMINO solution, while blue regions ($k_\text{DoMINO} < k_\text{lower} = 0.36 \rm\ m^2/s^2$) use the potential flow solution. The transition between the two solutions is smooth to avoid numerical artifacts.

Therefore, the ML-based initialization strategies can be summarized as:

\begin{itemize}
    \item \textbf{DoMINO + Uniform Flow (simple extension)}: DoMINO is used for the near-field, and uniform flow values are used for the far-field.
    \item \textbf{DoMINO (IDW extension)}: The point cloud from the DoMINO solution is combined with the points from the mesh's outer boundary (where the solution is known), and inverse distance weighting is used to interpolate to the full domain.
    \item \textbf{DoMINO + Potential Flow ($k$-based hybrid)}: The DoMINO solution is extended to the full domain using the IDW approach, and the DoMINO-predicted $k$ field is used to blend the DoMINO solution and potential flow solution.
\end{itemize}

\section{Results \& Discussion}

\subsection{Comparison of Initialization Strategies}

After 2 seconds of physical simulation time, all initializations had reached a statistically-steady state, with nearly-identical flow results.

\begin{figure}[htb!]
    \centering
    \includegraphics[width=\textwidth]{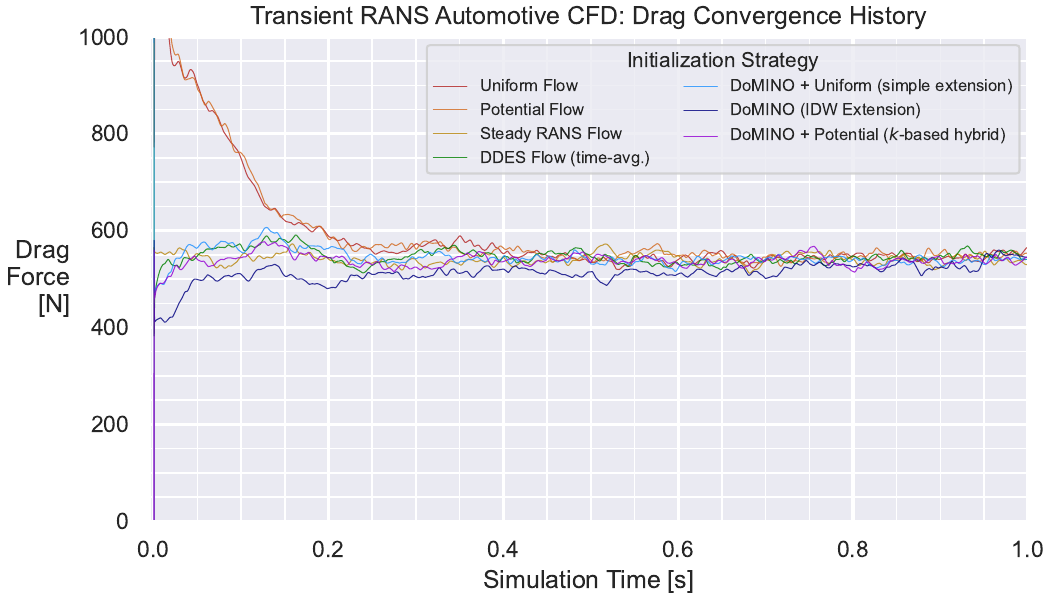}
    \caption{Predicted drag force over time for CFD simulations using different initialization strategies. See Figure \ref{fig:force_predictions_filtered} for a zoomed view.}
    \label{fig:force_predictions}
\end{figure}

However, the various initialization strategies had a significant impact on the time required to reach statistical steadiness, as defined by the metric later described in Section \ref{sec:convergence_metric}. In Figure \ref{fig:force_predictions}, we show the total drag force on the vehicle over time for cases using each of the initialization strategies described in Section \ref{sec:init_strategies}.

The simulations initialized with traditional inexpensive strategies (uniform flow, potential flow) exhibit a strong initial transient, with drag force errors exceeding 10\% relative to the final result for the first 0.2 seconds of simulation time (approximately 5.1 hours of wall-clock time). Even as far as 0.5 seconds into the simulation, Figure \ref{fig:force_predictions} shows that the drag force is still persistently higher than the final result with these initializations. In practice, errors of this magnitude mean that little useful time-averaging can begin, as the influence of the initial transient swamps the physical content of the solution.

In contrast, both the traditional expensive initializations (RANS and DDES-based) and the ML-based approaches (DoMINO with uniform flow extension and DoMINO with potential flow) provide meaningful initial predictions with much shorter initial transients. While these solutions still require time to fully converge, useful time-averaging can begin much earlier since the initial transient partially contains real physical content rather than numerical artifacts.

These results demonstrate that ML-based initializations can match the quality of traditional expensive approaches while dramatically reducing computational cost. RANS initialization required approximately 3 hours of wall-clock time to compute, and the DDES-based initializations from Ashton et al. \cite{ashtonDrivAerMLHighFidelityComputational2024} reportedly required 40 hours of wall-clock time on 1,536 cores. In contrast, the ML-based initialization with DoMINO requires about 5 seconds for inference on a 20 million cell mesh, and roughly 1 minute for interpolation to the simulation mesh.

\subsubsection{Implications for Initialization of Scale-Resolving Transient Methods}

An interesting observation in Table \ref{tab:convergence_times} is that the DDES-based initialization results in far slower convergence than the steady RANS initialization (12.9 hours vs. 4.7 hours). We suspect that is because the steady RANS initialization produces fields for the turbulence quantities of interest ($k$, $\omega$) using similar governing equations to the subsequent URANS solve, while the DDES-based study does not produce analogous data and hence initializes these turbulence quantities from uniform values. This evidence suggests that there is a significant loss of convergence speed associated with changes in turbulence modeling, and that $p$ and $\vec{U}$ initialization alone results in sub-par performance.

This finding has important implications when considering how the present findings might translate to ML-based initialization of a transient CFD solver that uses a DDES or LES method. In this case, a RANS initialization would provide information from a different turbulence modeling strategy, and hence may result in reduced performance. On the other hand, an existing ML model such as DoMINO could be trained directly on DDES or LES data with minimal modifications, and hence may provide more-transferrable initial guesses for the subsequent DDES or LES solver. If this extensibility to higher-fidelity methods is confirmed, this would enhance the industrial applicability of ML-based initialization methods; therefore, this is a high priority for further study.

\subsubsection{Industrial Implications}

The ML-based initializations demonstrate good generalization that transfers across different vehicle geometries, enhancing industrial applicability. This is demonstrated through the training process: the DoMINO model used here was trained on a different dataset of automotive geometries than the DrivAerML-derived test case, making this an out-of-distribution test case. The model's ability to provide useful initializations despite this suggests that the learned flow features transfer reasonably well across different vehicle geometries. This generalization capability is crucial for practical applications, as it means a single trained model can potentially provide initializations for a wide range of automotive designs. Furthermore, a workflow that uses ML surrogates for initialization alone allows a given model to be used in a wider range of cases than would be possible with a direct ML inference workflow, since the consequences of solution inaccuracies are mitigated by the subsequent transient CFD solve.

\subsection{Time-Averaging Procedure and Convergence Metric}
\label{sec:convergence_metric}

To more precisely quantify the computational time savings that are achievable using ML-based initializations, we developed a time-averaging procedure and statistical convergence metric. The time-averaging procedure was designed with several key requirements:
\begin{enumerate}
    \item It must only use backwards-looking data to enable real-time convergence assessment and termination,
    \item It should progressively downweight or forget older data to prevent initial nonstationary transients from contaminating the final stationary statistics,
    \item The effective sample size should increase over time to improve statistical confidence in the results.
\end{enumerate}

\begin{figure}[htb!]
    \centering
    \includegraphics[width=\textwidth]{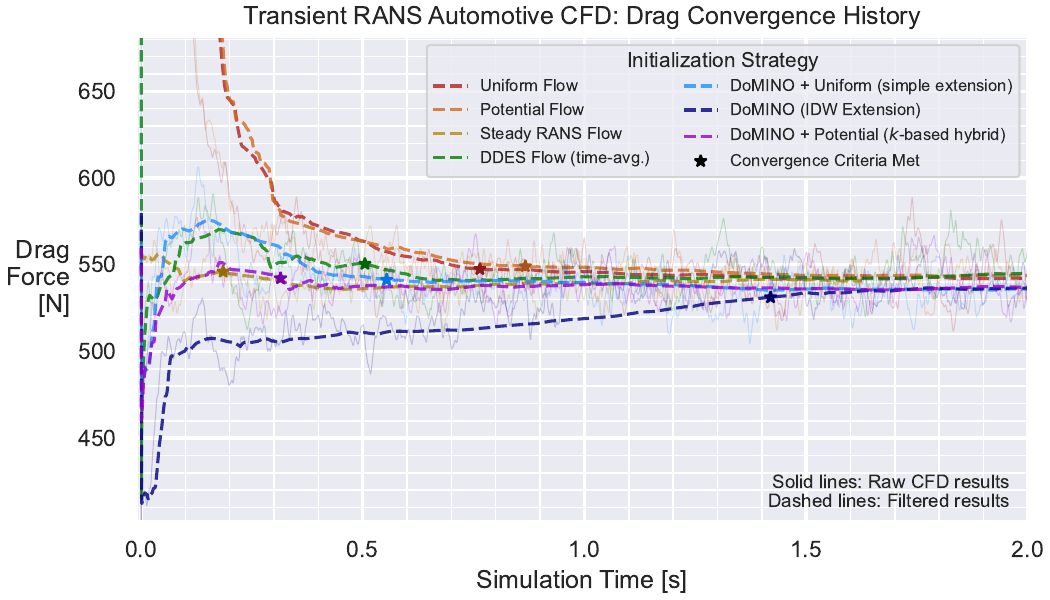}
    \caption{Predicted drag force over time for CFD simulations using different initialization strategies, with high-frequency noise removed using time-averaging. Zoomed view from Figure \ref{fig:force_predictions}.}
    \label{fig:force_predictions_filtered}
\end{figure}

To meet these requirements, we use a limited-window running median filter. At each timestep, the filter computes the median of the most recent 2/3 of the available data points. This fraction was chosen to balance statistical confidence (which improves with larger sample sizes) against the need to eventually forget initial transients. The results of this time-averaging procedure are shown in the dashed lines of Figure \ref{fig:force_predictions_filtered}. This gives a longer time window than Figure \ref{fig:force_predictions} but with a magnified force scale, to highlight the small-amplitude oscillations that occur during statistical convergence.

A forward-looking convergence metric is then computed based on the filtered results. Specifically, convergence is defined as the first point in time where the filtered result is within 1\% of its final value (at time $t=2 \rm\ s$) for all future times. This metric provides an objective way to compare the convergence times of different initialization strategies.

Figure \ref{fig:force_predictions_filtered} shows when each simulation has converged using these criteria, denoted by a star. These convergence times are also listed in Table \ref{tab:convergence_times}. Notably, the DoMINO + Potential Flow ($k$-based hybrid) strategy and the DoMINO + Uniform Flow strategy achieve statistical convergence in roughly half the time required using traditional inexpensive initializations (uniform flow, potential flow). In particular, the DoMINO + Potential Flow ($k$-based hybrid) results in convergence history that is very similar to that of a steady RANS flow, which is theoretically one of the closest-possible initializations for this case using traditional approaches.

\begin{table}[htb!]
    \centering
    \begin{tblr}{
            width=\textwidth,
            colspec={@{} X[0.1] X[2] X[1] X[1] X[1] @{}},
            row{1}={font=\bfseries},     % Bold header row
            column{1}={cmd=\textcolor{gray}},
            column{2}={font=\bfseries},     % Bold the initialization strategy
            row{3-Z}={m},                   % Vertically center data (except the header rows)
            cell{1}{1-3}={r=2}{}              % Merge row 1 and 2 for column 3
        }
        \toprule
                                                               & Initialization Strategy     & Initialization wall-clock runtime (hours) & \SetCell[c=2]{c,wd=4.2cm} {Time Required for Transient Convergence} &                            \\
        \cmidrule{4-5}
                                                               &                             &                                           & Physical simulation time (sec.)              & Wall-clock runtime (hours) \\
        \midrule
        \SetCell[r=4]{c} \rotatebox[origin=c]{90}{Traditional} & Uniform Flow                & Instant                                   & 0.7642                                       & 19.5                       \\
                                                               & Potential Flow              & 0.18                                      & 0.8668                                       & 22.1                       \\
                                                               & Steady RANS                 & 2.4                                       & 0.1852                                       & 4.7                        \\
                                                               & DDES Flow                   & 40$^\dag$                                 & 0.5050                                       & 12.9                       \\
        \midrule[lightgray]
        \SetCell[r=3]{c} \rotatebox[origin=c]{90}{ML-based}    & DoMINO + Uniform            & 0.02                                      & 0.5540                                       & 14.1                       \\
                                                               & DoMINO + IDW                & 0.03                                      & 1.4198                                       & 36.2                       \\
                                                               & DoMINO + Potential (hybrid) & 0.21                                      & 0.3146                                       & 8.0                        \\
        \bottomrule
    \end{tblr}
    \footnotesize\raggedright $^\dag$ As reported by Ashton et al. \cite{ashtonDrivAerMLHighFidelityComputational2024}, and run on different hardware (1,536 cores)
    \caption{Time required to reach statistical convergence for different initialization strategies, measured both in physical simulation time and equivalent wall-clock runtime. Listed wall-clock runtimes exclude the time required to compute the initialization itself. Unless marked, listed runtimes are measured on a 40-core dual-socket Intel Xeon E5-2698 v4 server.}
    \label{tab:convergence_times}
\end{table}

The convergence behavior in Figure \ref{fig:force_predictions_filtered} reveals another advantage of ML-based initializations: they provide meaningful physical content from the start. While traditional inexpensive methods show large-amplitude, non-physical transients in the first 0.2 seconds, both recommended ML-based strategies immediately begin capturing relevant flow physics. This allows useful time-averaging to begin earlier, even before formal convergence is achieved.

We observe that the DoMINO + IDW extension strategy achieves noticeably worse convergence than the other ML-based strategies. We believe that this is due to the fact that the IDW extension introduces large, systematic errors in the far-field region. In particular, the velocity ahead of the car is initialized to be roughly 5\% too low, which causes erroneously-low raw results until the correct velocity information from the inlet reaches the car, at approximately $t = (40\ \text{m}) / (38.889\ \text{m/s}) = 1.03\ \text{s}$. Interestingly, the direct cause of the force error is not the velocity error alone, but rather that this velocity error is not compensated by pressure error -- in other words, the \emph{total} pressure is the issue. Specifically, this results in a total pressure that is systematically too low, relative to the inlet boundary condition; this translates to lower stagnation pressure, and hence, erroneously-low drag force. While an error in \emph{static} pressure can be quickly corrected in a few iterations (as the information propagation speed for the pressure Poisson equation is only limited by numerics in incompressible flow), error in \emph{total} pressure must be physically advected out of the domain. The DoMINO + Uniform and DoMINO + Potential Flow strategies fundamentally avoid this issue, as their far-field values both provide a total pressure that is essentially identical to that of the final steady-state solution in the upstream region.

Because of this, the two recommended ML-based strategies for general use are the DoMINO + Uniform Flow and DoMINO + Potential Flow hybrid strategies. Nevertheless, the DoMINO + IDW strategy is useful to discuss, as it gives deeper insight into why certain kinds of initialization errors lead to poor convergence; this insight can be used to develop better ML-based initialization strategies.

\subsection{Flow Results}

Table \ref{tab:final_force} gives the final filtered value of the drag force for each initialization strategy, corresponding to the final value of the dashed lines in Figure \ref{fig:force_predictions_filtered}. These values agree closely across runs, although not exactly: the discrepancy between largest and smallest drag values is 1.8\% across the seven examined runs. These differences are of the same order of magnitude as the drag convergence tolerance of 1\%, and below the noise floor of typical expected drag accuracy with a $k$-$\omega$ SST URANS flow model on some practical cases\footnote{While expected accuracy is highly case-dependent, drag error against experiment for a $k$-$\omega$ SST URANS flow model on typical industrial external aerodynamics cases without extended laminar runs can be roughly $\pm$5\% in the grid-refined limit. \cite{morgadoXFOILVsCFD2016,adlerCFDNotCFD2022, roySummaryDataSixth2018}}. It is possible that with extended simulation runtime, all runs would asymptote to the same time-averaged drag value. 

On the other hand, there are documented cases in the literature observing hysteresis and bifurcation effects in both computational simulation and experiment \cite{clarkHighLiftPredictionWorkshop2025, hristovPostStallHysteresisFlow2017, huAerodynamicHysteresisLowReynoldsNumber2007}, particularly for flow cases with massive flow separation. Therefore, it is possible that the DoMINO-initialized near-field leads to transient convergence to a statistically-distinct flow field than would be the case with traditional initializations; determining whether this occurs is left as ongoing and future research. Notably, if this is determined to be the case, this information alone does not indicate whether either metastable state is more useful (i.e., representative of real-world flow scenarios) than the other -- further comparisons, ideally against wind tunnel data, would be required to assess this.

\begin{table}[htb!]
    \centering
    \begin{tblr}{
            width=\textwidth,
            colspec={@{} X[0.1] X[2] X[1.5] @{}},
            row{1}={font=\bfseries},     % Bold header row
            column{1}={cmd=\textcolor{gray}},
            column{2}={font=\bfseries},     % Bold the initialization strategy
            row{2-Z}={m},                   % Vertically center data (except the header rows)
        }
        \toprule
                                                               & Initialization Strategy     & Drag Force [N]  \\
        \midrule
        \SetCell[r=4]{c} \rotatebox[origin=c]{90}{Traditional} & Uniform Flow                & 542.0            \\
                                                               & Potential Flow              & 544.1            \\
                                                               & Steady RANS                 & 540.7            \\
                                                               & DDES Flow                   & 545.5            \\
        \midrule[lightgray]
        \SetCell[r=3]{c} \rotatebox[origin=c]{90}{ML-based}    & DoMINO + Uniform            & 535.5           \\
                                                               & DoMINO + IDW                & 536.5           \\
                                                               & DoMINO + Potential (hybrid) & 537.5           \\
        \bottomrule
    \end{tblr}
    \caption{Final filtered value of the drag force for different initialization strategies. Results agree closely, though slight differences that are on the same order of magnitude as the drag convergence tolerance (1\%) are observed.}
    \label{tab:final_force}
\end{table}

\begin{figure}[htb!]
    \centering
    \includegraphics[width=\textwidth]{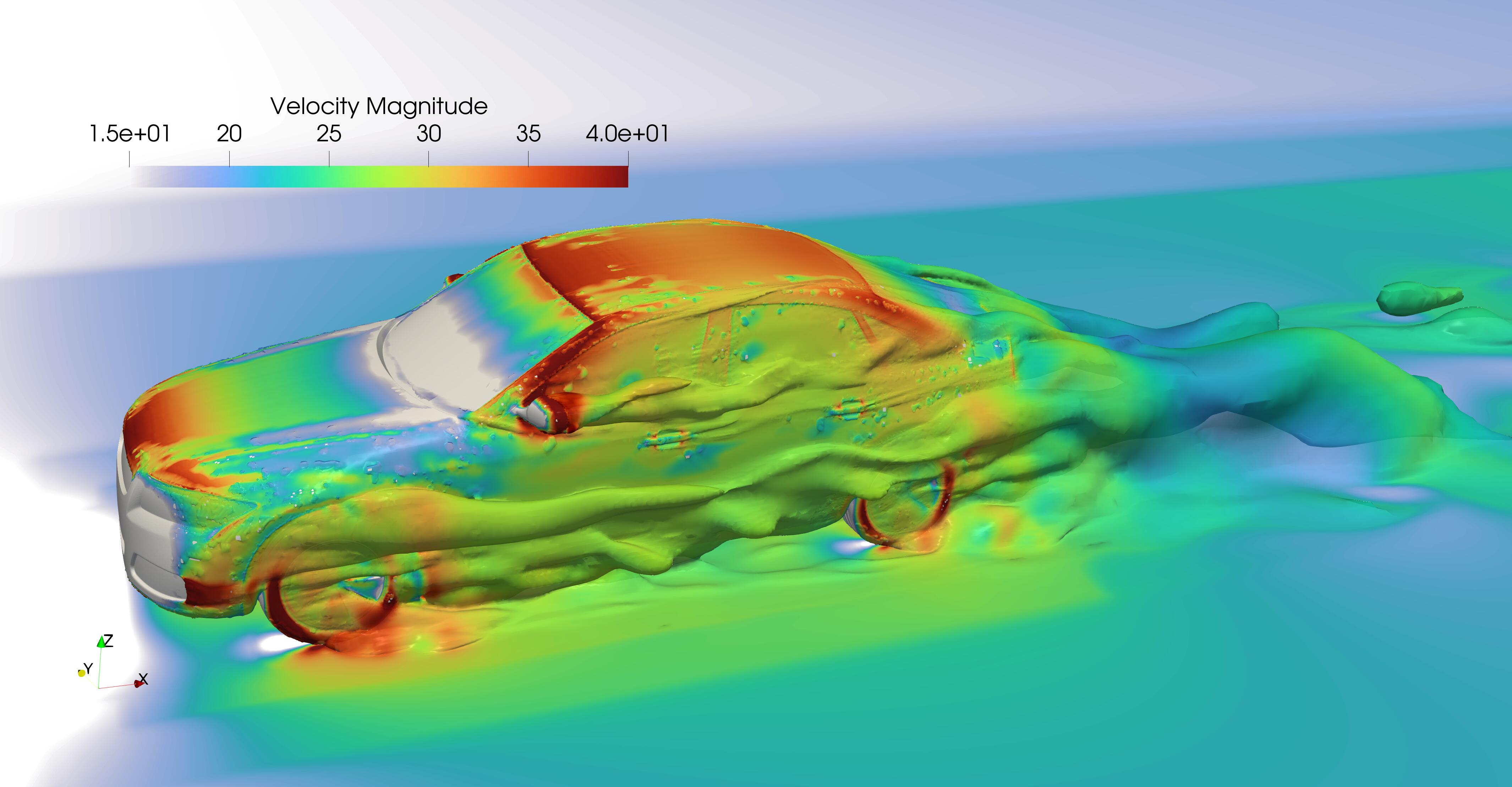}
    \caption{Total pressure isosurface showing turbulent structures in the wake region at $t = 2.0 \rm\ s$. Initialized with the DoMINO + Potential Flow ($k$-based hybrid) strategy.}
    \label{fig:ml_init_final_p}
\end{figure}

All initializations reached a statistically-steady state per the metric described in Section \ref{sec:convergence_metric}, with close agreement in overall flow fields by $t = 2.0 \rm\ s$. For illustration, the velocity field at the centerline plane in the simulation initialized with the DoMINO + Potential Flow ($k$-based hybrid) strategy is shown in Figure \ref{fig:ml_init_u_full} at $t = 2.0 \rm\ s$. The flow field exhibits the expected large-scale coherent turbulent structures, as shown in Figure \ref{fig:ml_init_final_p}. These structures are visualized using an isosurface of total pressure, which highlights regions of strong vortical motion in the wake of the vehicle. Notably, the wakes from the front wheel and mirror exhibit regular periodic oscillations, which illustrates that the first vortex shedding mode is well-resolved from these key features. This level of detail, where vortex shedding is resolved but subsequent turbulence cascade (i.e., vortex breakdown) are modeled, is typical of URANS simulations.

\begin{figure}[htb!]
    \centering
    \includegraphics[width=\textwidth]{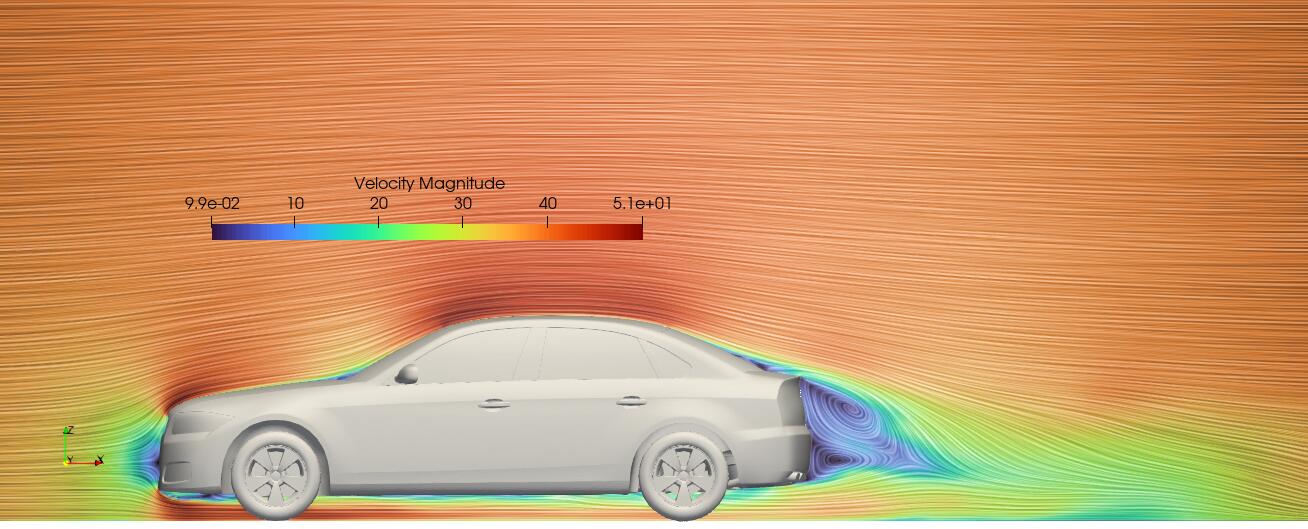}
    \caption{Velocity field in the simulation visualized on the centerline plane, at $t = 2.0 \rm\ s$ with statistically-steady flow. Initialized with the DoMINO + Potential Flow ($k$-based hybrid) strategy.}
    \label{fig:ml_init_final_u}
\end{figure}

\section{Conclusion}

In this work, we take first steps towards demonstrating that machine learning-based initializations can significantly accelerate the statistical convergence of transient CFD simulations, relative to common existing methods. Using a case study in automotive aerodynamics, we showed that ML-based initialization strategies can reduce the time required to reach statistical steadiness by approximately 50\% compared to traditional inexpensive approaches like uniform or potential flow initializations. This improvement brings the convergence time in line with that achieved using computationally expensive initializations (e.g., steady RANS), but at a fraction of the computational cost.

The hybrid approach combining DoMINO predictions with potential flow solutions proved particularly effective on this external aerodynamics case. By leveraging DoMINO's accuracy in high-vorticity regions while retaining potential flow's exact satisfaction of conservation laws in the far-field, this strategy achieved rapid convergence while mitigating the numerical artifacts that can arise from pure ML-based approaches. Notably, this performance was achieved despite the DoMINO model being trained on a different dataset of automotive geometries, demonstrating a degree of generalization capability that is crucial for practical applications.

Another notable initialization strategy explored in this work was the DoMINO + Uniform Flow approach, which achieved comparable performance to the potential flow hybrid while making fewer assumptions about the underlying physics. While the potential flow hybrid strategy demonstrated excellent performance for automotive aerodynamics, its success relies on assumptions specific to external aerodynamics problems -- namely, that the flow has large regions with zero vorticity, and that these regions can be consistently identified. This assumption, while valid for thin shear layer flows around streamlined bodies, may not hold for other classes of physics problems (e.g., internal duct flows). In contrast, the DoMINO + Uniform Flow strategy makes no such physical assumptions, as it simply uses ML predictions where they are expected to be accurate and falls back to a simple uniform state otherwise. This physics-agnostic approach suggests broader applicability across different types of transient CFD simulations. This flexibility, combined with the strong convergence performance demonstrated in our automotive test case, suggests that the DoMINO + Uniform Flow approach may be particularly valuable for developing general-purpose initialization strategies that work across multiple physics domains. While specialized approaches like the potential flow hybrid may offer further improvements for specific applications, the broader applicability of the uniform flow strategy could make it a more practical choice for industrial workflows that must handle diverse simulation types.

These results suggest a promising path forward for industrial CFD workflows. Traditional approaches have forced practitioners to choose between computationally expensive but accurate initializations (like steady RANS) and inexpensive but less-accurate and potentially-destabilizing alternatives (like uniform or potential flow). ML-based initializations offer a compelling third option: reasonably-accurate initial fields that can be rapidly generated. This capability is particularly valuable in industrial settings where multiple design iterations may need to be evaluated, as the reduced time to statistical convergence directly translates to increased throughput in the design process.

Future work could explore the extension of these techniques from URANS methods to more computationally expensive transient solvers, such as large-eddy simulation (LES) and delayed detached-eddy simulation (DDES) methods. Because the time to statistical convergence in both URANS and higher-fidelity methods is fundamentally limited by physical advection timescales rather than numerical considerations, we hypothesize that a similar \emph{relative} speedup could be achieved for these methods. If so, this could translate to higher \emph{absolute} wall-clock runtime savings, due to the typically-higher per-iteration cost of LES and DDES methods. As the automotive industry increasingly adopts LES and DDES methods for aerodynamics analysis, this could translate to significant additional practical benefits.

Beyond solver types, future work could also explore other classes of flow problems, particularly those where traditional initialization strategies struggle to provide meaningful initial conditions (e.g., aeroacoustics, propeller-wing interactions, combustion, etc.). Additionally, the development of ML architectures specifically designed for initialization (rather than final flow prediction), and the tuning and optimization of various intermediate steps (such as the ML model's inference resolution, or mesh interpolation methods) could potentially yield further improvements in convergence time.

% bibliography
\bibliographystyle{plain}
\bibliography{C:/Users/psharpe/library, references}

\end{document}